\documentclass[11pt,a4paper]{article}
\usepackage[hyperref]{acl2019}
\usepackage{times}
\usepackage{latexsym}

\usepackage[shortlabels]{enumitem}
\usepackage{url}
\usepackage{subfiles}
\usepackage{amsmath}
\usepackage{bm}
\usepackage{booktabs}
\usepackage{float}
\usepackage{smartdiagram}
\usepackage[justification=centering]{caption}
\usepackage{subcaption}
\usepackage{graphicx}
\usepackage{tikz}
\usetikzlibrary{positioning}
\usetikzlibrary{decorations.pathreplacing}
\usesmartdiagramlibrary{additions}
\usepackage[utf8x]{inputenc}
\def\dis{\displaystyle}

\aclfinalcopy 

\renewcommand\vec{\boldsymbol}

\graphicspath{{./}}

\DeclareMathOperator*{\argmax}{argmax}

\title{One-Shot Learning for Language Modelling}

\author{%
  Talip Uçar \\
  \texttt{ucabtuc@ucl.ac.uk} \\
  \And
  Adrian Gonzalez-Martin \\
  \texttt{adrian.martin.18@ucl.ac.uk} \\
  \AND
  Matthew Lee \\
  \texttt{matthew.lee.16@ucl.ac.uk} \\
  \And
  Adrian Daniel Szwarc \\
  \texttt{adrian.szwarc.18@ucl.ac.uk} \\
}

\date{}

\begin{document}
\maketitle
\begin{abstract}
    Humans can infer a great deal about the meaning of a word, using the syntax and semantics of surrounding words even if it is their first time reading or hearing it. We can also generalise the learned concept of the word to new tasks. Despite great progress in achieving human-level performance in certain tasks \cite{Silver2016}, learning from one or few examples remains a key challenge in machine learning, and has not thoroughly been explored in Natural Language Processing (NLP). In this work we tackle the problem of one-shot learning for an NLP task by employing ideas from recent developments in machine learning: embeddings, attention mechanisms (softmax) and similarity measures (cosine, Euclidean, Poincare, and Minkowski). We adapt the framework suggested in matching networks \cite{VinyalsMatch}, and explore the effectiveness of the aforementioned methods in one, two and three-shot learning problems on the task of predicting missing word explored in \cite{VinyalsMatch} by using the WikiText-2 dataset.
    
    Our work contributes in two ways: Our first contribution is that we explore the effectiveness of different distance metrics on k-shot learning, and show that there is no single best distance metric for k-shot learning, which challenges common belief. We found that the performance of a distance metric depends on the number of shots used during training. The second contribution of our work is that we establish a benchmark for one, two, and three-shot learning on a language task with a publicly available dataset that can be used to benchmark against in future research. 
\end{abstract}

\section{Introduction}\label{sec:introduction}

In the last few years, there has been great progress in the applications of deep neural networks, especially in speech \cite{Hinton2012}, computer vision (CV) \cite{Krizhevsky2012} and NLP \cite{Mikolov2010}. Their success has mostly relied on the availability of large datasets and powerful hardware (TPU, GPU). Despite their success, these networks are still not good at generalising concepts from few samples, and require re-training on new data and for new tasks. This is mainly due to the parametric nature of these models, in which one needs large datasets to generalise better. However, non-parametric alternatives such as k-Nearest Neighbour (k-NN) avoid this problem, and incorporate new data better i.e. one can use a k-NN to classify a new class without having to do any optimisation, but its performance depends on the chosen metric such as  L2 distance \cite{Atkeson1997}. One solution would be training a fully end-to-end nearest neighbour classifier.

Hence, the shortcomings of the current state of the art neural networks and possible advantages of non-parametric models make us wonder: Can we do better? Can we build networks that can learn from few examples? This question motivates “one-shot” learning through training a fully end-to-end nearest neighbour classifier. We define one-shot learning as learning a class from a single labelled example. Wed use the Matching Networks of \cite{VinyalsMatch} as our basic model to incorporate
different similarity measures, embeddings, and attention mechanisms to improve generalisation from few examples.

Our contribution is twofold: at the modelling level, and at the benchmarking level. Firstly, we explore new similarity metrics in the context of one-shot learning and how they match up with different embedding schemes. Secondly, due to the exploratory nature of our work, we believe that it can be used to benchmark other approaches on language modelling using WikiText-2 dataset.

\begin{figure*}
    \begin{center}
    \includegraphics[keepaspectratio, width=0.75\paperwidth]{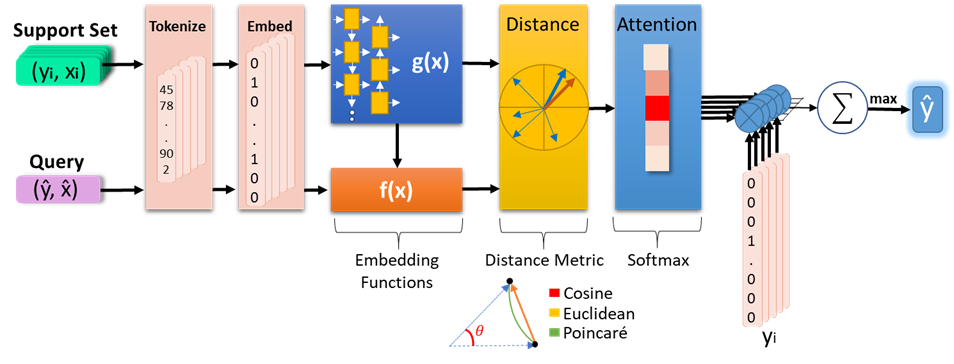}
   \caption{\textbf{Architecture:} Embedding + Distance Metric + Attention }
    \end{center}
\end{figure*}

We have organised this report by first defining and explaining our architecture in Section 2. Then, in Section 3, we briefly discuss some of the related work to the task and our model. In Section 4, we describe  our setup and the experiments we performed, demonstrating strong results in one-shot learning for language modelling in matching the old baseline set by \cite{VinyalsMatch}. In Section 5, we present and analyse our findings. Finally, in Section 6, we conclude with a summary of our research outcomes, as well as future research directions one might take.

\section{Model \& Methods}\label{sec:methods}

Our approach to one-shot learning has multiple aspects. First, our architecture follows matching networks \cite{VinyalsMatch}, in which we use embedding layers to encode sentences, distance metrics to measure similarity, and an attention scheme that uses softmax. Secondly, our training approach is different than traditional supervised learning routines. One-shot learning is easier if the network is specifically trained to do one-shot learning. Therefore, we want the training-time protocol to be exactly the same as the test-time protocol. In this case, given a support set, S (N classes with k examples each, where k=1,2, or 3), the model matches new samples to one of N classes.

\subsection{Model Architecture}
We describe our model as a differentiable nearest neighbor. This analogy becomes clear through the equation that we use to compute the target $\hat{y}$ for a test sample $\hat{x}$:

    \begin{equation}
        \hat{y} = \sum \limits_{i=1}^k a(\hat{x}, x_i) y_i
    \end{equation}

\noindent where we are assigning label of $\hat{x}$, based on its similarity to training examples $x_1,\ldots, x_k$ by using kernel \textbf{a(.)} to get a weighted average of labels $y_i$ of the training examples. In this case, $x_i$ represents the embedding of a sentence, and $y_i$ is a one-hot vector of the size of the vocabulary.

\begin{figure*}
    \begin{center}
    \includegraphics[keepaspectratio, width=0.6\paperwidth]{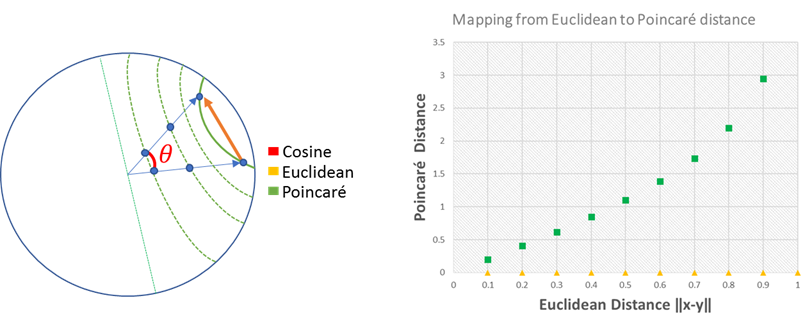}
   \caption{\textbf{Comparing distance metrics:} Cosine, Euclidean, and Poincaré}
    \end{center}
\end{figure*}

\subsection{Embedding \& Attention}

We embed both the training examples $x_i$ and the test example $\hat{x}$. We use a distance metric such as cosine similarity to decide how similar the test sample is to each of the training samples. We pass these similarity measurements through a softmax to get normalized weights:

\begin{equation}
        a(\hat{x}, x_i)
        = \frac{e^{c(f(\hat{x}), g(x_i))}}
               {\sum_{j=1}^k e^{c(f(\hat{x}), g(x_j))}}
        \label{eq:attention-mechanism}
    \end{equation}

\noindent where $c()$ is the cosine distance, but we will later discuss using other distance metrics. The important thing to note here is that we compute the embedding using both the test sample and the support set S together so that the embedding is learned through the relationship between the test and training samples.

\subsubsection{Pre-Embedding}
We have two-level embeddings. In the first level of embedding, or pre-embedding, we embed each sentence using max-pooling sentence embedding. This establishes our baseline. But, we might want to replace max-pooling with any other embedding method such as BERT, or wrod2vec in a future work. Since each method introduces differently sized embedded vectors, we standardise the interface between the first and second level embedding by using a linear network to reduce the dimension of the sentence embedding to 64 just before the second level embedding. The second level embedding is known as a Full Context Embedding (FCE) as was introduced by \cite{VinyalsMatch}.

\subsubsection{Full Context Embedding}

Figure-1 shows two embedding functions $f(x)$ and $g(x)$. Function $f(x)$  is used to embed the test sample into a vector while $g(x)$ embeds each training example into a vector. $g(x)$ is a bidirectional LSTM over the examples:

    \begin{align}
        g(x_i, S) =  \overrightarrow{h}_{i} + \overleftarrow{h}_{i} + g'(x_i) 
    \end{align}
    \begin{align}
        \overrightarrow{h}_{i}, \overrightarrow{c}_{i} &= \texttt{LSTM}(g^\prime(x_i), \overrightarrow{h}_{i-1}, \overrightarrow{c}_{i-1}) 
    \end{align}
    \begin{align}
        \overleftarrow{h}_{i}, \overleftarrow{c}_{i} &= \texttt{LSTM}(g^\prime(x_i), \overleftarrow{h}_{i-1}, \overleftarrow{c}_{i-1})
    \end{align}
    
\noindent where the encoding of the i'th example $x_i$ is a function of its pre-embedding $g'(x_i)$ ($g'$ might seem confusing but is common in the literature) and the embedding of other examples in set S. The relationship between examples is built through the bidirectional network's hidden states. In this way, each training example becomes a function of itself as well as other samples in the set. This particular feature is what separates the Matching Network from typical nearest neighbour methods since nearest neighbour does not change the representation of a sample based on other data points in the training set. Also, it is important to note that since each sample in set S is randomly picked, there is no particular order in which we feed them to the bidirectonal LSTMs. So, it is possible that we would get different embeddings if we permute the examples, which is something left to do as a future work.

 $f(x)$ is used to embed the test example. We use support set, S, to affect the encoding of test sample, so $f(\hat{x})$ is actually $f(\hat{x}, S)$ i.e. a function of test and training samples. Specifically, $f(\hat{x}, S)$ is a vanilla LSTM with an attention mechanism that unfolds over K steps:
    
    \begin{equation}
        f(\hat{x}, S) = \texttt{attLSTM}(f^\prime(\hat{x}), g(S), K)
    \end{equation}
 
 It processes K-time steps to attend over the examples in the training set. The encoding is the last hidden state of the LSTM:

    \begin{align}
        & \hat{h}_k, c_k = \texttt{LSTM}(f^\prime(\hat{x}), [h_{k-1}, r_{k-1}], c_{k-1}) \\
        & h_k = \hat{h}_k + f^\prime(\hat{x})\\
        & r_{k} = \sum \limits_{i=1}^{|S|} a(h_{k-1}, g(x_i))g(x_i)\\
        & a(h_{k-1}, g(x_i)) = \texttt{softmax}(h_{k-1}^T g(x_i))
    \end{align}

 We should note that the hidden state is a function of the previous hidden state as well as a vector r, which is computed by attending over the encoded training examples i.e. the output of the aforementioned g(x) function. In this way, the encoding of the test sample becomes a function of the training samples.

\begin{figure*}
    \begin{center}
    \includegraphics[keepaspectratio, width=0.6\paperwidth]{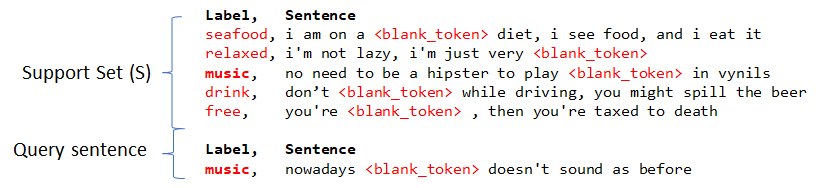}
   \caption{\textbf{One-shot learning task}}
    \end{center}
    \label{fig:language-modelling-task}
\end{figure*}

\subsection{Distance Metrics}

The original work of \cite{VinyalsMatch} uses the cosine distance as a similarity measures, and does not explore other possibilities. Since we are using a kernel as a weighting on labels, performance of this model partly depends on the choice of distance metric, though see the results and discussion. Hence, we explored four possible options among others to compare their effect on performance: cosine, Euclidean, Poincaré, and Minkowski (p=1,3).

Figure-2 shows what each distance metric means in terms of relating two embeddings. The cosine distance finds the cosine of the angle between the two embeddings so it is thus a judgement of orientation and not magnitude: two vectors with the same orientation have a cosine similarity of 1 while orthogonal vectors have a similarity of 0, regardless of their magnitude.
However, the Euclidean distance between two points is the length of the path connecting them. The Minkowski distance is a generalised metric form of Euclidean distance and Manhattan distance. Different metrics arise from the Minkowski general formula:

$$\left(\sum_{i=1}^n |x_i-y_i|^p\right)^{1/p}$$ where

\begin{itemize}
    \item p = 1 is Manhattan distance, L1-Norm
    \item p = 2 is Euclidean distance, L2-Norm
    \item p = ∞ is Chebyshev distance, Lmax-Norm
\end{itemize}

We are using p=1, and 3 in our experiments. Lastly, the Poincaré distance is shown to be successful in predicting links in graphs where they outperform Euclidean embeddings in terms of representation capacity and in terms of generalisation ability \cite{Poincare17}. The Poincaré distance between points $u, v \in B^d$ is:

    \begin{equation}
        \dis d(\mathbf{v},\mathbf{u}) = \texttt{arcosh}(1+2\frac{\lVert \mathbf{u-v} \rVert^2 }{(1-\lVert \mathbf{u} \rVert^2 )(1-\lVert \mathbf{v} \rVert^2 ) })
    \end{equation}
    
\noindent where $B^d = \{x \in R^d | \lVert \mathbf{x} \rVert < 1 \} $ is a d-dimensional unit ball, the Poincaré ball. We can see that the distance within the Poincaré ball changes smoothly with respect to the location of $u$ and $v$. We observed that, during training, this smoothness manifested itself as loss going down more smoothly (i.e. less spiky) than other distance metrics and there was less variance on the validation error between batches. Moreover, \cite{Krioukov10} modelled complex networks using hyperbolic spaces and discussed how typical properties such as strong clustering emerges by assuming an underlying hyperbolic geometry. This was appealing to us since our underlying model can be considered a clustering method. 

\begin{table*}[h]
  \centering

  \begin{tabular}{@{}llll@{}}
    \toprule
       & \textbf{Model-based}
       & \textbf{Metric-based}
       & \textbf{Optimization-based} \\
    \midrule
        \textbf{Key idea}
            & RNN, memory
            & Embeddings, distance metric
            & Training optimization \\
        \textbf{How is $P(\vec{y} \mid \vec{x})$ modeled?}
            & $f(\vec{y}, \mathcal{S})$
            & $\sum_{i} k(\vec{x}, \vec{x}_{i})\vec{y}_{i}$
            & $P(\vec{y} \mid \vec{x})$ \\
    \bottomrule
  \end{tabular}
  
  \caption{Summary of different approaches to Few-Shot Learning}
  \label{table:common-approaches}
\end{table*}

\subsection{Training}

We want the test-time protocol to exactly match the training time protocol. Our task is as follows: given N classes (words) with k sentences each (k = 1,2 or 3), predict a new label from one of N classes. We do this in two steps:

\begin{itemize}
    \item Sample a task T from the training data by selecting N labels (N=5, 20), and k sentences per label (k=1,2, or 3).
    \item Generate an episode by sampling a label set L (i.e. {"music", "dogs"...}) and then use L to sample the support set S and a batch B of examples to evaluate the loss on.
\end{itemize}

Let $\theta$ be the parameters of the model, and $P_\theta$ a probability distribution. Then the matching net is trained to minimise the error predicting $P$'s labels, conditioned on $S$ with the following training objective:
    \begin{equation}
        \theta = \argmax_\theta \textcolor{red}{ E_{L \sim T} } \left[  E_{ \textcolor{red}{S \sim L} , P \sim L} \left[K \right] \right]
    \end{equation}
    \noindent,\text{where }
    
    \begin{equation}  
                  K = \sum \limits_{(x,y) \in P} \log P_\theta(y \mid x, \textcolor{red}{S}) 
     \end{equation}          

\noindent where symbols in red are added for meta-learning in addition to the supervised learning objective. Figure-3 shows examples of training and test samples for a 5-way, one-shot learning task. Meta-learning is another term for the training and testing paradigm used here and was discussed in more detail in our previous report.

\section{Related Work}\label{sec:literature-review}

With the recent resurgence of deep learning, a new body of work on one-shot, and few-shot learning has started to emerge. There are three main approaches to one-shot, and few-shot learning: 

\begin{enumerate}[label=\textbf{\alph*})] 
  \item \textbf{Metric-based}. Focuses on distance metrics \cite{Koch15, VinyalsMatch, SnellProto, Sung18}
  
  \item \textbf{Model-based}. Uses recurrent networks with attention as well as external or internal memory \cite{Santoro16, Mun17}
  
  \item \textbf{Optimization-based}. Optimizes the model parameters explicitly \cite{Ravi2017, Finn17, Nic18}. 
\end{enumerate}

\noindent Table~\ref{table:common-approaches} summarises these observations.

Our work is mostly influenced by metric-based models, which combine ideas such as attention, and embeddings.

\cite{Koch15} proposed a method to use the Siamese Neural Network (SNN) to do one-shot learning in vision. The model is trained to predict whether two input images are in the same class. During test time, it processes all the image pairs between a test image and every image in the support set, and makes a prediction using the class of the support image with the highest probability. One of the shortcomings of this work is that it heavily depends on a pre-trained model, whose advantage decreases when the new task diverges from the original task that the model was trained on.

\cite{Sung18} proposed Relation Networks, which is similar to SNN but with a few differences: Firstly, the relationship between images is captured by a CNN classifier instead of using the L1 norm distance metric. Secondly, they use the mean-square-error loss as the objective function instead of the cross-entropy since their network focuses on predicting relation scores rather than binary classification.

The Prototypical Networks of \cite{SnellProto} use some of the same insights as in Matching Networks \cite{VinyalsMatch} and simplifies Matching Networks. One major difference between the two approaches is the distance metric:  Prototypical Networks uses the Euclidean distance instead of the cosine similarity. However, both are equivalent in the case of one-shot learning since they can be reduced to k-nearest neighbours (k-NN).

Although each aforementioned work offers a new idea, they are mostly designed for vision tasks. Among them, only the Matching Network has tried to solve a language task, and made an attempt to establish a benchmark, though not on a publically available dataset as we do. Moreover, each fails to explore and compare the effectiveness of other distance metrics. Lastly, since their publication, there have been new developments on word and sentence embeddings such as BERT, and Poincaré embeddings \cite{Poincare17}, which enable researchers to build more expressive language models.

\section{Experiments}\label{sec:experiments}

\begin{table*}[h]
  \centering

  \begin{tabular}{@{}llllll@{}}
    \toprule
        \multicolumn{1}{c|}{\textbf{}} 
        & \multicolumn{3}{c|}{5 Class}  
        & \multicolumn{2}{c}{20 Class} \\ 
        \textbf{Distance Metric}
        & \textbf{1-Shot}
        & \textbf{2-Shot}
        & \textbf{3-Shot}
        & \textbf{1-Shot}
        & \textbf{5-Shot} \\
    \midrule
        \textbf{Cosine}
            & $28.6\%$
            & $32.8\%$
            & $34.1\%$
            & $10.7\%$ 
            & $14.7\%$ \\
        \textbf{Euclidean}
            & $30.1\%$
            & $31.0\%$
            & $35.4\%$\\
        \textbf{Poincare}
            & $28.1\%$
            & $30.6\%$
            & $35\%$\\
        \textbf{Minkowski (p=1)}
            & $27.5\%$
            & $30.5\%$
            & \bm{$37.7\%$}\\
        \textbf{Minkowski (p=3)}
            & $29.1\%$
            & $32.0\%$
            & $35.5\%$ \\
        \textbf{Matching Networks}
            & \bm{$32.4 \pm 0.2\%$}
            & \bm{$36.1 \pm 0.1\%$}
            & \bm{$38.2 \pm 0.2\%$} \\
            
    \bottomrule
  \end{tabular}
  
  \caption{Accuracy results of different distance metrics on One, Two, and Three-Shot models for the WikiText-2 dataset. `FCE` enabled in all cases, and all are trained with `5` processing steps on the `attLSTM` layer. The final row in bold is for the results for the same task in \cite{VinyalsMatch}.}
  \label{table:accuracy-results-review}
\end{table*}

For our experiments, we define an N-way k-shot learning task as following: Given k (1,2, or 3) labelled examples for a support set of sentences (N=5, or 20), in which each has a missing word and a corresponding 1-hot label, and is not previously trained on, choose the label from the support set that best matches the query sentence. Sentences are taken from the WikiText-2 dataset.

On each trial, the support set and batches are populated with non-overlapping sentences. Thus, we do not use infrequent words since if there is only a single sentence for a given word it would not work since the sentence would need to be in both the set and the batch. 

Each trial consisted of a 5-way choice between the classes available in the set. We used a batch size of 20 throughout the sentence matching task and varied the set size across k=1,2, and 3. We made sure that the same number of sentences were available for each class in the set. We split the words into a randomly sampled 9000 for training, 1000 for cross-validation and 1000 for testing. Our results are based on the test set. Thus, words and sentences used during test time were not trained on during training.

\textbf{Baseline:}  
We use Matching Networks as our baseline model, in which we use:
\begin{itemize}
    \item Max-pooling sentence embedding
    \item Cosine similarity
    \item Full context embedding enabled
\end{itemize}

We build on this baseline by exploring different distance metrics. Table~\ref{table:accuracy-results-review}  compares results for our experiments and the original Matching Network.

\section{Results and Discussion}\label{sec:results-discussion}

In this section, we present the results of our experiments with similarity metrics and embeddings. As mentioned before, all the experiments followed the pattern of an $N$-way $k$-shot learning task.

Table~\ref{table:accuracy-results-review} shows the accuracy of trials with different distance metrics, and compares against results of the original Matching Network \cite{VinyalsMatch}. Our first experiment was to reproduce the results of the original Matching Networks with a simple embedding scheme. Compared to the original paper, we managed to achieve very similar accuracy in each of the tests with results ranging from $30.1\%$ for $5$-class $1$-shot learning to $37.7 \%$ for 5-class 3-shot. 
Ju

Exploration of different metrics introduced to our model has shown some insights. To list a few, we will use 5-Class k-shot learning results:

\begin{itemize}
    \item Euclidean distance did better than Cosine across in 1, and 3-shot learning, while performed worse in 2-shot learning. This challenged the findings of Prototypical Networks \cite{SnellProto}, where the claim was that Euclidean was much better as a similarity measure.
    \item Accuracy of all metrics increases with increased number of samples (i.e. shots).
    \item Although Euclidean distance outperformed others in 1-shot learning, Cosine was best in 2-shot learning. The biggest performance gain is achieved using Minkowski (p=1) i.e. Manhattan distance in 3-shot learning. This is somewhat surprising since it shows that there is no single best metric, and the performance of a distance metric depends on number of examples (shots) used. This challenges some of the claims made by other authors, where the claim was superiority of one distance metric over others.
    \item Poincare distance did not perform well. This poor performance is mainly due to the basic embedding scheme we used. We believe that if we used embeddings learned with Poincare distance this would boost its performance. This remains as future work.
\end{itemize}

 Although the accuracy achieved for the 5-Class $k$-shot task was very positive in terms of matching past research, we have noticed very low accuracy for all the models performing 20-class $k$-shot learning. As a result we have completely dropped the 20-way $k$-shot learning task and focused on 5-way $k$-shot. This also made us realise why \cite{VinyalsMatch} most likely did not report findings using more than 5 classes.

\begin{figure}
  \includegraphics[width=\linewidth]{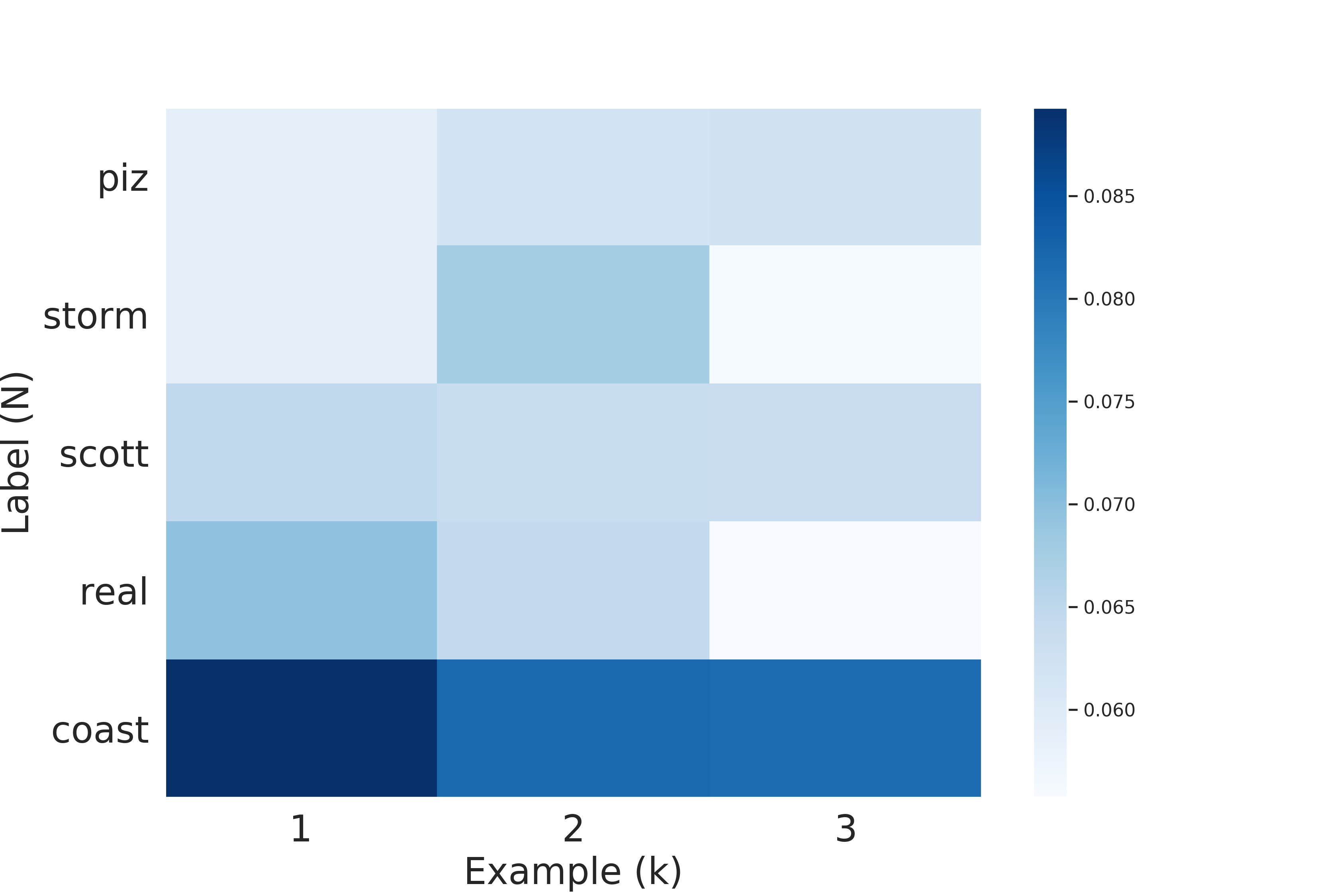}
  \caption{Attention map for target label $coast$ with Euclidean distance}
  \label{fig:attention-map}
\end{figure}

To further investigate the similarity performance we have plotted an attention map for the support set and target label (See figure~\ref{fig:attention-map}). In the plot the $x$-axis was assigned to the $k$ number of example sentences and the $y$-axis to the $N$ number of labels in support set. Although Euclidean distance was not the best choice of similarity metrics for 5-way 3-shot learning, it can be clearly seen that the model performs matching correctly. The target word was $coast$ and it was accurately allocated with the highest attention to all three example sentences from support set.    

\begin{figure}
  \includegraphics[width=\linewidth]{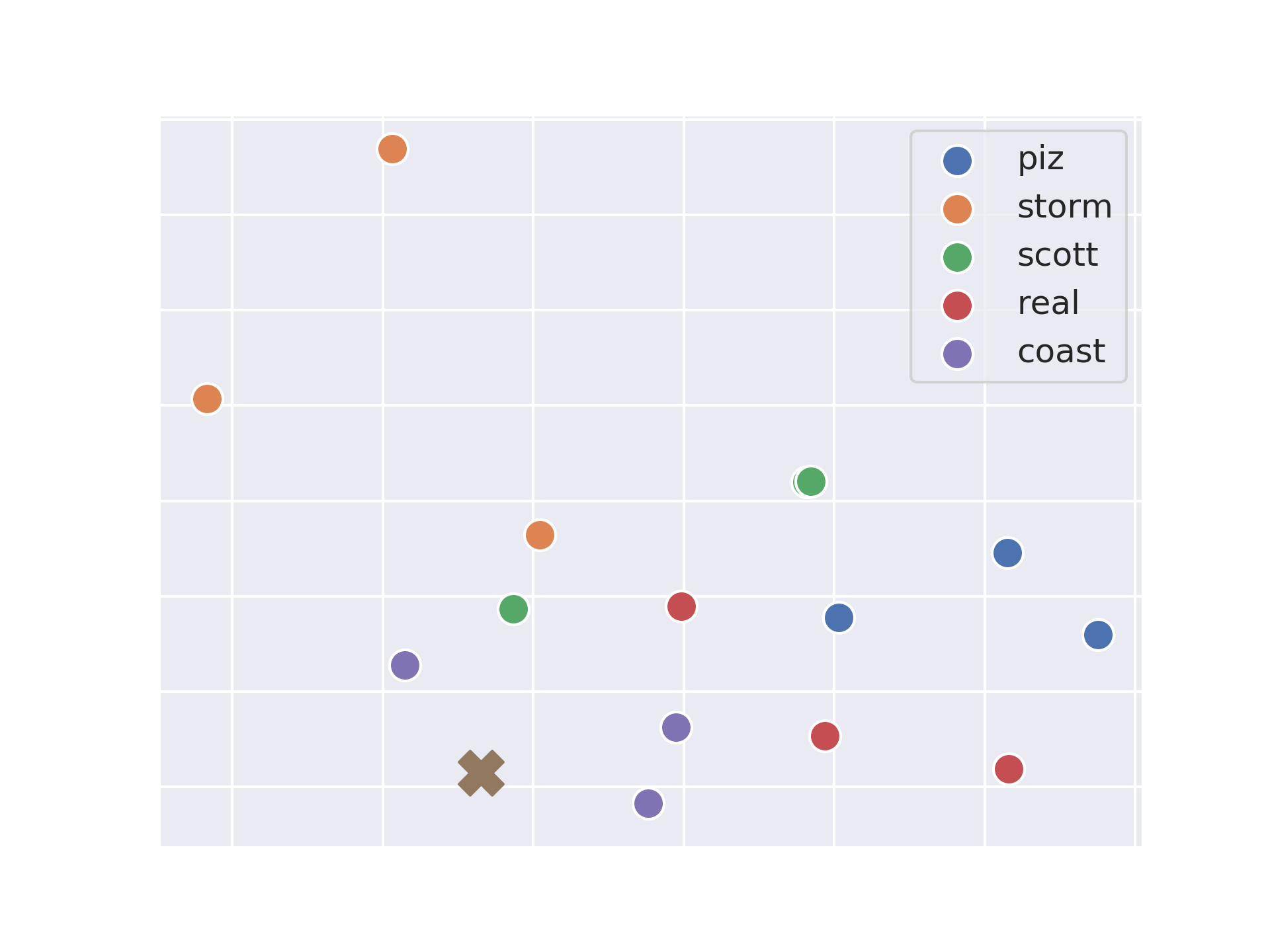}
  \caption{Clustering of example labels' embeddings in relation to $coast$ target label}
  \label{fig:emb-plot}
\end{figure}

 Moreover, in Figure~\ref{fig:emb-plot}, it can be observed that the correct example labels have been clustered near the target label proving accurate performance of the embedding process between target and support sets. However, one issue noticed during the experiments was related to embedding different examples of the same label. To illustrate, in figure~\ref{fig:emb-plot}, some examples of the same support set label are not appropriately clustered, as the distances between them should be relatively short (for instance $storm$ label examples). The overall accuracy of the process could perhaps be improved by introducing a different embedding, for instance BERT \cite{Bert}, and word2vec. We did generate Bert embeddings at the end of our project but ran out of time to finish training.

\textbf{A few concerns:}
During our work, we came to realise a few weaknesses of our approach:

\begin{itemize}
    \item In bidirectional LSTMs, we feed sets of training examples to encode them, but, the order is random. Depending on which order we choose, we get a different encoding for the same support set. This could be resolved by using a recurrent attentional mechanism instead \cite{Vinyals2015}
    
    \item This approach also gets quite a bit slower as the number of training examples grow. So we think that for larger datasets, parameteric approaches could be more suitable for this task.
    
    \item Another weakness of this approach is that we use a specific number of examples, e.g. 5-20 during training, which forces us to use the same number at test time. This is problematic if we want the size of our training set to grow online. This would require retraining the network because the encoder LSTM for the training data is not "used to" seeing inputs of more examples. This could be resolved by iteratively sub-sampling the training data, doing multiple inference passes and averaging. Moreover, the attention mechanism LSTM might have problems with attending over many more examples. An interesting experiment to try would be to disable FCE and try to use up to 1000 training examples, and compare it with the case where we train on up to 25 examples (with and without FCE). 
    \item Another candidate baseline that we could have chosen to compare to would be SVM, but this is left as a future work.
    \item Performance of each distance metric was limited by the fact that we used relatively simple max-pooling embedding. And we feel that we maxed out the performance. To do better, one needs to used more expressive embeddings such as BERT. Also, we believe that the best performance would be gained through using sentence embedding that was acquired using same similarity measure that is being used to compare distances. In our case, we can consider max-pooling embedding as being more friendly to Euclidean measures, but not so much to Poincare. One of future direction in this work is to explore Poincare further by using Poincare embeddings.

\end{itemize}

\section{Conclusion}\label{sec:conclusion}

We explored the effectiveness of different embedding schemes and distance metrics in one-shot learning for a language modelling task. We used Matching Networks, a differentiable nearest neighbor model, as our starting point, and trained it end-to-end. We contributed towards improving one-shot learning problem in an NLP task by exploring distance metrics, and establishing a benchmark for future research. What remains to be done in near future is exploring the same problem by using more powerful sentence embeddings, and achieving state-of-the art result in this task. Our code can be found in this GitHub repository \footnote{https://github.com/adriangonz/statistical-nlp-17}

\section*{Acknowledgments}

We would like to thank Oriol Vinyals for his helpful discussions on his paper, "Matching Networks for One Shot Learning" and giving us insights and suggestions. We also thank Sebastian Ruder for giving us helpful directions on one-shot learning, and Patrick Lewis for feedback on our project. This project idea was developed from reading \cite{Ruder18}. We also benefited from writings of Lilian Weng on one-shot learning \cite{lillog18}.

\bibliography{main}

\bibliographystyle{acl_natbib}

\end{document}